\documentclass[10pt,twocolumn,letterpaper]{article}

\usepackage{iccv}
\usepackage{times}
\usepackage{epsfig}
\usepackage{graphicx}
\usepackage{amsmath}
\usepackage{amssymb}
\usepackage{booktabs}
\usepackage{pifont}
\usepackage{appendix}
\usepackage[accsupp]{axessibility}  

\usepackage[pagebackref=true,breaklinks=true,letterpaper=true,colorlinks,bookmarks=false]{hyperref}

\iccvfinalcopy 

\def\iccvPaperID{2624} 

\ificcvfinal\fi

\begin{document}

\title{EmoTalk: Speech-Driven Emotional Disentanglement for 3D Face Animation}

\author{Ziqiao Peng$^1$\quad Haoyu Wu$^1$\quad Zhenbo Song$^2$\quad Hao Xu$^{3,6}$\quad Xiangyu Zhu$^4$\\ Jun He$^{1}$\quad Hongyan Liu$^{5*}$\quad Zhaoxin Fan$^{1,6}\thanks{corresponding authors}$ \\
$^1$Renmin University of China\quad $^2$Nanjing University of Science and Technology\\
$^3$The Hong Kong University of Science and Technology\quad $^4$Chinese Academy of Sciences \\
$^5$Tsinghua University\quad $^6$Psyche AI Inc. \\
{\tt\small \{pengziqiao, wuhaoyu556, hejun, fanzhaoxin\}@ruc.edu.cn\quad songzb@njust.edu.cn}\\
{\tt\small hxubl@connect.ust.hk\quad xiangyu.zhu@nlpr.ia.ac.cn\quad liuhy@sem.tsinghua.edu.cn}
\and
}

\maketitle
\ificcvfinal\thispagestyle{empty}\fi

\begin{abstract}
Speech-driven 3D face animation aims to generate realistic facial expressions that match the speech content and emotion. However, existing methods often neglect emotional facial expressions or fail to disentangle them from speech content. To address this issue, this paper proposes an end-to-end neural network to disentangle different emotions in speech so as to generate rich 3D facial expressions. Specifically, we introduce the emotion disentangling encoder (EDE) to disentangle the emotion and content in the speech by cross-reconstructed speech signals with different emotion labels. Then an emotion-guided feature fusion decoder is employed to generate a 3D talking face with enhanced emotion. The decoder is driven by the disentangled identity, emotional, and content embeddings so as to generate controllable personal and emotional styles. Finally, considering the scarcity of the 3D emotional talking face data, we resort to the supervision of facial blendshapes, which enables the reconstruction of plausible 3D faces from 2D emotional data, and contribute a large-scale 3D emotional talking face dataset (3D-ETF) to train the network. Our experiments and user studies demonstrate that our approach outperforms state-of-the-art methods and exhibits more diverse facial movements. We recommend watching the supplementary video: \url{https://ziqiaopeng.github.io/emotalk}

\end{abstract}

\section{Introduction}

Dynamic and realistic speech-driven facial animation has garnered growing interest in virtual reality~\cite{wohlgenannt2020virtual,fan2022object,fan2022deep}, computer gaming~\cite{ping2013computer,fan2022reconstruction,boyle2014narrative}, and film production~\cite{liu2009analysis,ye2022perceiving,chen2023msp}. For current commercial products, 3D face blendshape is handcrafted by animators, whereas manual scripts drive facial expressions. Such a process demands substantial expenses and considerable time and labor. As deep learning techniques are utilized in various scenarios~\cite{pouyanfar2018survey}, deep end-to-end speech-driven facial animation~\cite{karras2017audio,cudeiro2019capture,richard2021MeshTalk,fan2022faceformer,chen2022transformer} has been widely studied in industry and academia. Presently, learning-based 3D facial animations can not only produce high-quality animation effects but also facilitate cost reduction during production.

\begin{figure}[t]
\begin{center}
   \includegraphics[width=.9\linewidth]{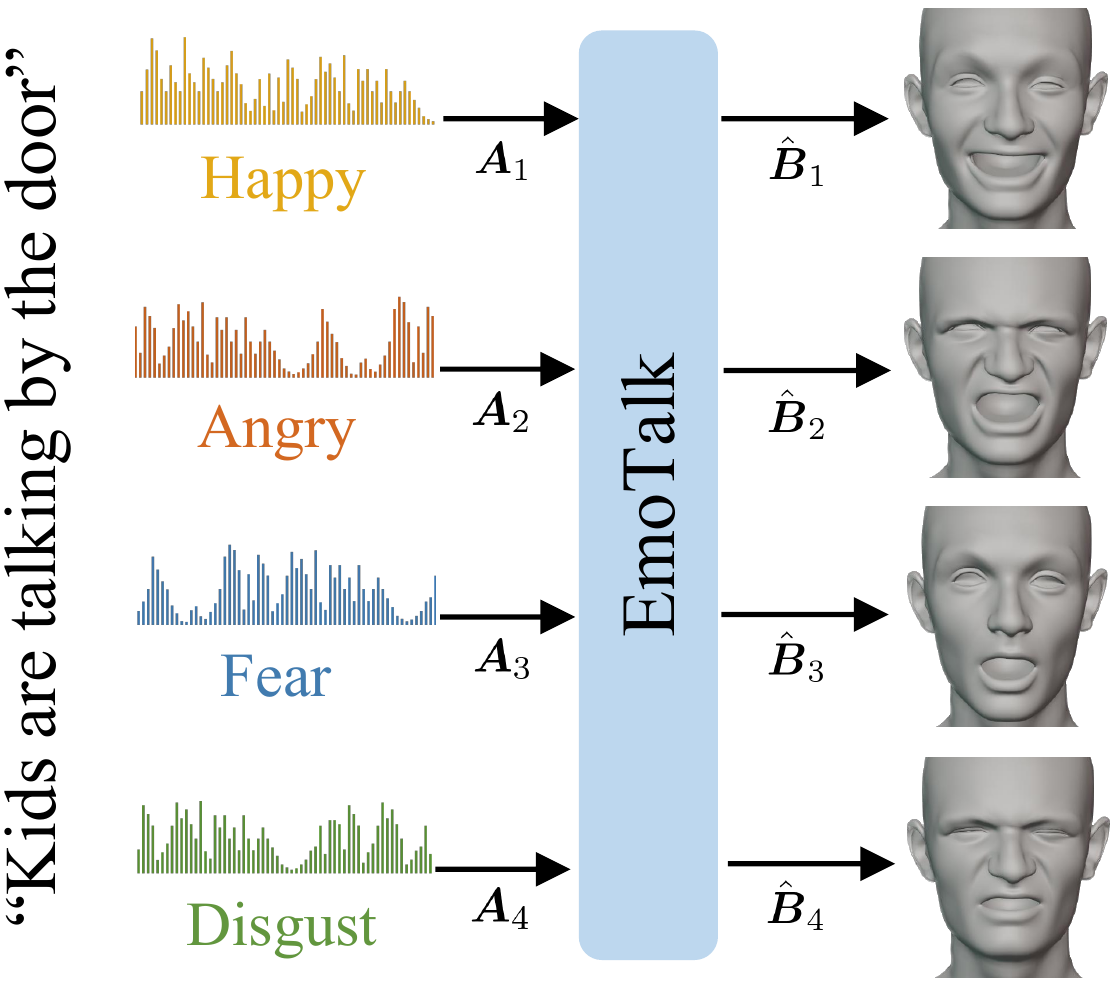}
\end{center}
   \caption{\textbf{Results of EmoTalk.} Given audio input expressing different emotions, EmoTalk produces realistic 3D facial animation sequences with corresponding emotional expressions as outputs.}
\label{fig:1}
\end{figure}

However, current methods mainly focus on improving the synchronization between lip movements and speech~\cite{sheng2022deep}, neglecting the emotional variation of facial expressions. We argue that emotions are an essential aspect of human communication and expression, and emotion absence in 3D facial animations may cause the uncanny valley effect. It is a crucial issue to recover emotional expressions for the speech-driven 3D face animation problem. In fact, emotional information is naturally contained in the speech, and extracting emotions is a crucial task for speech understanding~\cite{vogt2008automatic}. Nevertheless, as audio content and emotion are entangled, it is hard to extract explicit content and emotion from a speech simultaneously. In order to generate rich emotional facial expressions, previous 2D facial animation methods encode the emotions manually and only learn the content feature from the speech~\cite{papantoniou2022neural,chang2022disentangling,sun2022continuously}. By manipulating the emotion code, the facial decoder could achieve appropriate emotional modulation. Manually controlling may generate changeable emotions, but it could result in contradiction with the emotion in speech. For example, it does not conform to human intuition by inputting angry speech but outputting a happy expression. \par

To address this issue, we propose a novel speech-driven emotion-enhanced 3D facial animation method (Fig. \ref{fig:1}) in this paper, where an emotion disentangling encoder and emotion-guided feature fusion decoder are proposed to consist of our key contribution, as illustrated in Fig.~\ref{fig:2}. For the emotion disentangling encoder, two distinct audio feature extractors~\cite{baevski2020wav2vec} are introduced and utilized to extract two separate latent spaces for the content and emotion, respectively, which is exploited to decouple emotion and content. A cross-reconstruction loss is further presented to constrain the learning process to better disentangle the emotion and content from the speech. While for the emotion-guided feature fusion decoder, multiple different types of features are decoded by a Transformer~\cite{vaswani2017attention} module with periodic positional encoding and emotion-guided multi-head attention, which will output 52 emotion-enhanced blendshape coefficients to represent the final human facial expressions. Extensive experiments show that our method significantly outperforms current state-of-the-art methods in terms of emotional expression by disentangling content and emotion.

To train the proposed network, emotional speeches with corresponding 3D facial expressions are required. However, as far as we know, there is no publicly available 3D emotional talking face dataset that we can use, posing a serious new challenge.
To tackle the issue, a large-scale pseudo-3D emotional talking face dataset, termed the 3D-ETF dataset, is further introduced in our work. To build this dataset and make it more applicable, we first collaborated with several professional animators to create 52 FLAME head templates~\cite{li2017learning} that are semantic meaningful. Then, ``pseudo" 3D blendshape labels are generated from images of large-scale audio-visual datasets~\cite{livingstone2018ryerson,zhang2021flow} by utilizing a well-established 3D facial blendshape capture system. Finally, the 3D-ETF dataset with both blendshape coefficients~\cite{lewis2014practice} and mesh vertices are constructed through blend linear skinning. Since its blendshape labels are semantic meaningful, the 3D-ETF dataset is versatile, allowing the facile transfer of facial movements among different virtual characters~\cite{pawaskar2013expression}. \par

In summary, the main contributions of our work are as follows:
\begin{itemize}
\item[$\bullet$] We propose an end-to-end neural network for speech-driven emotion-enhanced 3D facial animation, which achieves various emotional expressions and outperforms existing state-of-the-art methods.
\item[$\bullet$] We introduce the emotion disentangling encoder, which disentangles the emotion and content in the speech and makes the facial animation aware of clear emotional information.
\item[$\bullet$] We present a large-scale 3D emotional talking face (3D-ETF) dataset including both blendshape coefficients and mesh vertices. We have implemented parameterized transformations for blendshape coefficients and the FLAME model, allowing for efficient conversion between various facial animations.

\end{itemize}

\section{Related Work}

\subsection{Speech-driven 3D facial animation}
Previously, numerous studies have been conducted on 2D talking head generation~\cite{chen2019hierarchical,zakharov2019few,mittal2020animating,guo2021ad,zhang2021facial,thies2020neural,ji2021audio,hong2022depth}, which uses image-driven or speech-driven approaches to create realistic videos of speaking individuals. However, these methods are not applicable to 3D character models that are widely used in 3D games and virtual reality interactions. Therefore, speech-driven 3D facial animation has attracted more attention recently~\cite{cao2005expressive,karras2017audio,cudeiro2019capture,tian2019audio2face,hussen2020modality,richard2021MeshTalk,liu2021geometry,fan2022faceformer,chen2022transformer}. \par

\begin{figure*}
\begin{center}
    \includegraphics[width=0.99\textwidth, keepaspectratio]{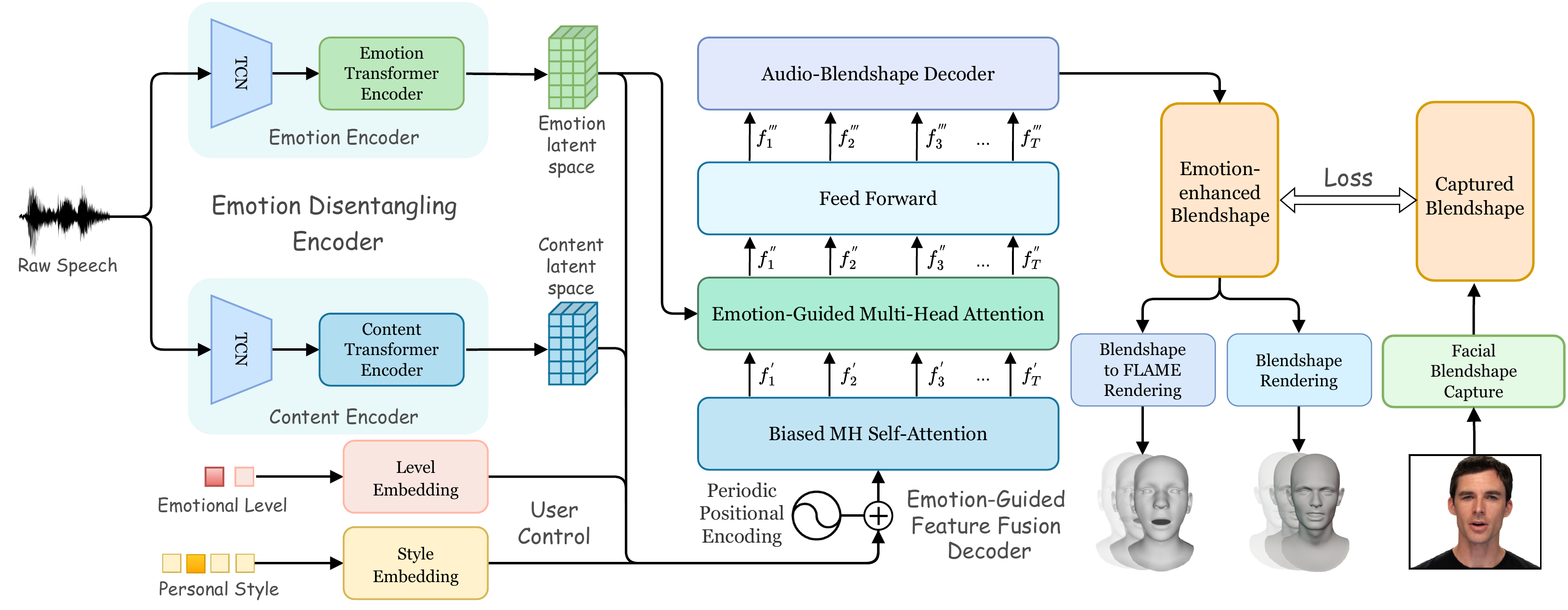}
\end{center}
   \caption{\textbf{Overview of EmoTalk.} Given a speech input $\boldsymbol{A}_{1:T}$, emotional level $\boldsymbol{l}$, and personal style $\boldsymbol{p}$ as inputs, our model disentangles the emotion and content in the speech using two latent spaces. The features extracted from these latent spaces are combined and fed into the emotion-guided feature fusion decoder, which outputs emotion-enhanced blendshape coefficients. These coefficients can be used to animate a FLAME model or rendered as an image sequence.}
\label{fig:2}
\end{figure*}
One of the challenges in this field is the lack of high-quality datasets for various emotions. For example, VOCA~\cite{cudeiro2019capture} uses time convolutions and control parameters to generate realistic character animation from any speech signal and a static character mesh. Still, it only produces decent mouth movements due to the limited upper face movement in the VOCASET dataset~\cite{cudeiro2019capture}. Similarly, FaceFormer~\cite{fan2022faceformer} uses a Transformer-based model to obtain contextually relevant audio information and generates continuous facial movements in an autoregressive manner. It improves multi-source generalization and achieves more precise changes in mouth movements than VOCA but still does not enhance facial expressions because it also uses the VOCASET dataset.\par
MeshTalk~\cite{richard2021MeshTalk} focuses on the upper part of the face, which is lacking in VOCA, and creates a categorical latent space for facial animation, disentangles audio-correlated and audio-uncorrelated movements through cross-modality loss, thereby synthesizing audio-uncorrelated movements such as blinking and eyebrow movements. Although MeshTalk has achieved upper face movement, the current methods still need to solve the problem of lack of emotion in 3D facial animation due to the absence of emotional facial animation datasets.\par

\subsection{Speech emotion recognition and disentanglement}
Speech emotion recognition (SER) is an essential but challenging task for generating realistic talking head animations. Various techniques have been proposed in the paper to extract emotions from speech signals, such as traditional speech analysis and classification methods ~\cite{nwe2003speech,schuller2003hidden,kwon2003emotion,huang2014speech,schuller2018speech,hossain2019emotion,jain2020speech}. In this paper, we focus on deep learning-based SER techniques to learn features from speech signals. For example, Mekruksavanich \etal~\cite{mekruksavanich2020negative} used one-dimensional CNNs~\cite{lecun1998gradient} to achieve 96.60$\%$ accuracy in classifying negative emotions from a Thai language dataset. Yenigalla \etal~\cite{yenigalla2018speech} combined phoneme sequences and spectrograms as inputs to a CNN model and obtained better SER performance than using either input alone.

One of the key challenges in speech processing is to separate emotion from content, which enables the neural network to learn more specific features. This process is called emotion disentanglement. Several methods exist for achieving emotion disentanglement and reconstruction in speech using different techniques, such as variational autoencoding Wasserstein generative adversarial networks (VAW-GANs) ~\cite{9383526}, cross-speaker emotion transfer~\cite{zhang2022iemotts}, and Mel Frequency Cepstral Coefficient (MFCC)~\cite{muda2010voice} with Dynamic Time Warping (DTW)~\cite{muller2007dynamic}. In this paper, we build on the work of Ji \etal~\cite{ji2021audio}, who decomposed speech signals into two decoupled spaces and fed them into a facial synthesis module. We propose an improved emotion disentangling encoder for 3D facial animation generation, which will be described in detail in Sec. \ref{section:cross}.

\section{Method}
We propose a 3D facial animation model that can reconstruct facial expressions with rich emotions from speech signals, enabling users to control emotional level and personal style. Let $\boldsymbol{A}_{1: T}=\left(\boldsymbol{a}_{1}, \ldots, \boldsymbol{a}_{T}\right)$ be a sequence of speech snippets, and each $ \boldsymbol{a}_{t} \in \mathbb{R}^{D}$ has $\boldsymbol{D}$ samples to align to the corresponding (visual) frame $\boldsymbol{b}_{t}$. Let $\boldsymbol{B}_{1: T}=\left(\boldsymbol{b}_{1}, \ldots, \boldsymbol{b}_{T}\right), \boldsymbol{b}_{t} \in \mathbb{R}^{52}$ be a $\boldsymbol{T}-$length sequence of face blendshape coefficients, and each frame is represented by 52 values. The whole pipeline of our approach is revealed in Fig. \ref{fig:2}. By analyzing emotional information from any arbitrary speech signal $\boldsymbol{A}_{1: T}$, our method is capable of producing differentiated face coefficients $\hat{\boldsymbol{B}}_{1: T}$. Moreover, the proposed model takes a user-controllable emotional level $\boldsymbol{l} \in \mathbb{R}^{2}$ as input, which allows users to modulate the strength of the expressed emotions in the resulting facial animations. Personal style $\boldsymbol{p} \in \mathbb{R}^{24}$ inputs can also be manipulated by users to have different speaking habits. These two parameters are the same one-hot encoding as~\cite{tran2017disentangled}.
Then, the decoder predicts facial coefficients $\hat{\boldsymbol{B}}_{1:T}=\left(\hat{\boldsymbol{b}}_{1}, \ldots, \hat{\boldsymbol{b}}_{T}\right)$ conditioned on speech representations $\boldsymbol{A}_{1: T}$, the emotional level $\boldsymbol{l}$, and the personal style $\boldsymbol{p}$. Formally, 
\begin{equation}\label{1}
\boldsymbol{\hat{b}}_{t}=\mathrm{EmoTalk}_{\theta}(\boldsymbol{a}_{t},\boldsymbol{l},\boldsymbol{p}),
\end{equation}
where $\theta$ indicates the model parameters. For the convenience of describing detailed network components, let $\boldsymbol{A}_{ci,ej}$ denote the sample data pertaining to the $\boldsymbol{i}^{th}$ content and $\boldsymbol{j}^{th}$ emotion in the audio sample, whereas $\boldsymbol{B}_{ci,ej}$ denote the sample data pertaining to the $\boldsymbol{i}^{th}$ content and $\boldsymbol{j}^{th}$ emotion in the blendshape coefficients sample. Both representations will be employed in the following sections to introduce the details of our method. \par

\subsection{Emotion disentangling encoder}
\label{section:cross}

The intricate relationship between speech and facial expressions makes it arduous to learn the mapping from speech to emotional facial expressions directly. To address this issue, we propose an improved emotion disentangling encoder for 3D facial animation generation, inspired by Ji \etal~\cite{ji2021audio}. To the best of our knowledge, this is the first work that applies emotion disentanglement to this task. Our module simplifies and enhances the original disentanglement module in several ways. First, we replace the MFCC~\cite{muda2010voice} feature extractor, which cannot capture rich speech information and has a complex input process, with a pre-trained audio feature extractor wav2vec 2.0~\cite{baevski2020wav2vec}. Second, we streamline the disentanglement process to enhance its conciseness and comprehensibility. Third, we transform the module into an end-to-end form that directly outputs 52 blendshape coefficients required for facial animation, allowing the model to receive better constraints during training.\par

\begin{figure}[t]
\begin{center}
   \includegraphics[width=1\linewidth]{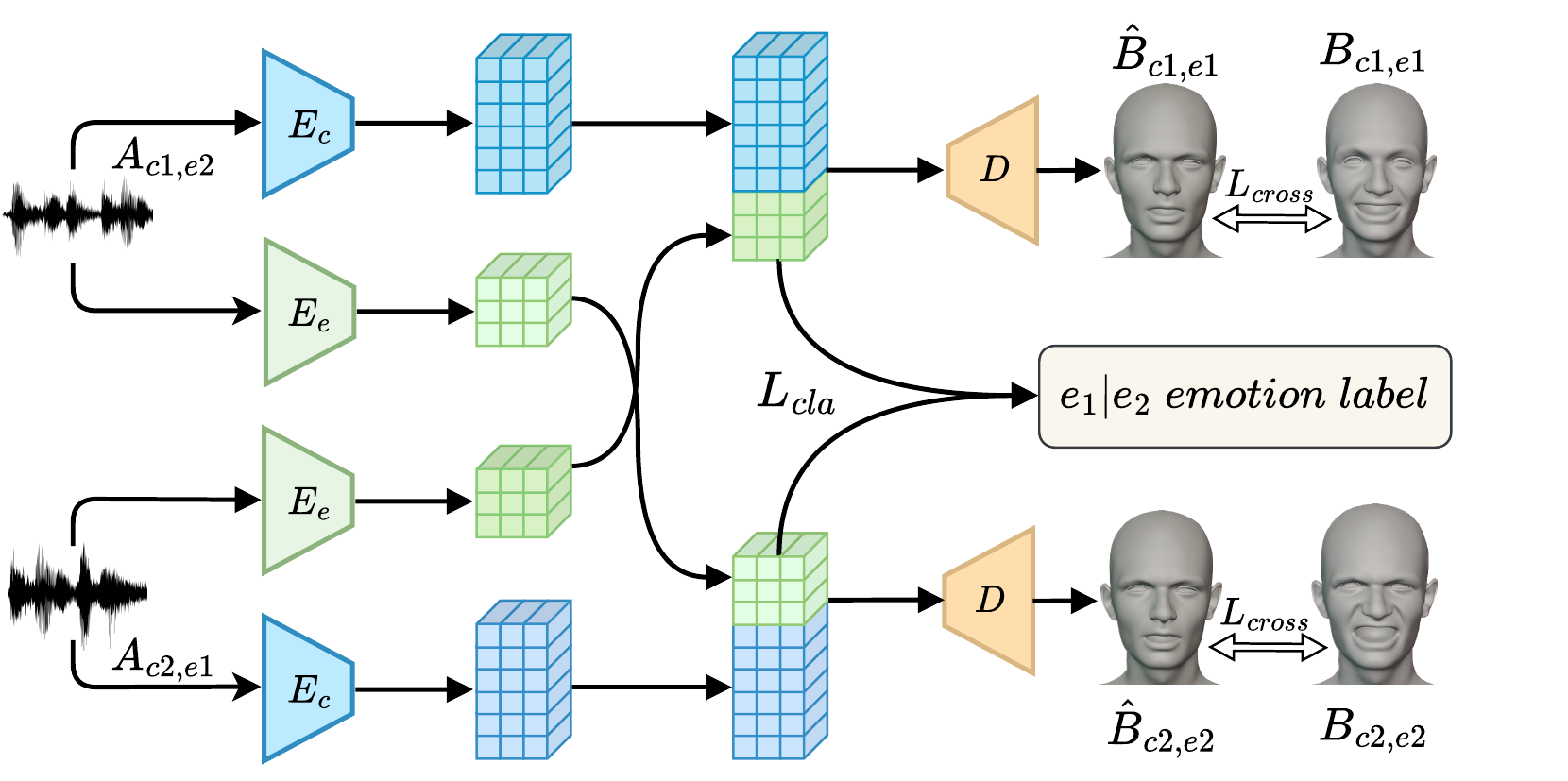}
\end{center}
   \caption{\textbf{Emotion Disentangling Encoder.} Various inputs of speech, conveying different contents and emotions, are processed to generate cross-reconstructed blendshape coefficients representing distinct combinations of facial expressions.}
\label{fig:3}
\end{figure}
\noindent\textbf{Reorganization and disentanglement.} As illustrated in Fig. \ref{fig:3}, the emotion disentangling encoder is designed to disentangle short-term content features from long-term emotion features in speech. 

Nevertheless, the module cannot guarantee the disentanglement between content and emotion. To achieve this objective, we utilize pseudo-training pairs that combine diverse emotions and contents as input and require the network to reconstruct the corresponding ground truth samples as output. This approach compels the network to acquire disentangled content and emotion representations, which can better capture both aspects of speech and enhance the overall performance of the model.\par

To separate content and emotion features in speech, two pre-trained audio models~\cite{baevski2020wav2vec} are used as feature extractors $\boldsymbol{E}_{c}$ and $\boldsymbol{E}_{e} \in \mathbb{R}^{1024}$, which are fine-tuned on content and emotion, respectively. The pre-trained models' temporal convolutional network(TCN) layer~\cite{lea2016temporal} is fixed during fine-tuning since it is trained on a considerable amount of audio data. We input two audios $\boldsymbol{A}_{c1,e2}$ and $\boldsymbol{A}_{c2,e1}$, where the subscript $\boldsymbol{c}$ denotes text content, and $\boldsymbol{e}$ denotes audio emotion. Content features $\boldsymbol{c1}$ and $\boldsymbol{c2}$ are extracted using $\boldsymbol{E}_{c} \left( \boldsymbol{A}_{c1,e2} \right) $ and $\boldsymbol{E}_{c} \left( \boldsymbol{A}_{c2,e1} \right) $, respectively. Emotion features $\boldsymbol{e1}$ and $\boldsymbol{e2}$ are extracted using $\boldsymbol{E}_{e} \left( \boldsymbol{A}_{c2,e1} \right) $ and $\boldsymbol{E}_{e} \left( \boldsymbol{A}_{c1,e2} \right) $, respectively. The content and emotion features are concatenated and fed into a decoder module that outputs face blendshape coefficients for reconstruction. Pseudo-training pairs comprising different combinations of content and emotion are used as input, and the network is required to reconstruct the corresponding ground truth samples as output, namely $\boldsymbol{\hat{B}}_{c1,e1}$ and $\boldsymbol{\hat{B}}_{c2,e2}$, for constraints with real samples $\boldsymbol{B}_{c1,e1}$ and $\boldsymbol{B}_{c2,e2}$. This approach enforces disentanglement between content and emotion features by requiring that they can be combined to reproduce both aspects of speech.\par

\subsection{Emotion-guided feature fusion decoder}
In this work, we propose an emotion-guided feature fusion decoder that maps audio to 3D facial animation coefficients using emotional information from audio. This approach aims to generate more expressive facial animations. This module consists of four components: emotion features $ \boldsymbol{F}_{e} \in \mathbb{R}^{256}$ and content features $ \boldsymbol{F}_{c} \in \mathbb{R}^{512}$ extracted from two latent spaces, personal style features $ \boldsymbol{F}_{p} \in \mathbb{R}^{32}$ that control the individual characteristics of facial expressions, and emotional level features $ \boldsymbol{F}_{l} \in \mathbb{R}^{32}$ that regulate the degree of emotional expression. These four features are concatenated along the same dimension and subsequently inputted into the emotion-guided feature fusion decoder.

To generate the 3D blendshape coefficients from the fused feature, we employ a module similar to the Transformer~\cite{vaswani2017attention} decoder. The input feature $\boldsymbol{F}$ is first encoded with periodic positional encoding~\cite{fan2022faceformer}, which captures the stable open and close times of lip movements during speech. Then, a biased multi-head self-attention layer that integrates positional encoding into multi-head attention layers inspired by attention with linear biases (ALiBi)~\cite{press2021train} produces $\boldsymbol{f}_{t}^{'}$, which assigns higher weights to closer information in the mask layer and focuses on the changes between adjacent actions. Subsequently, an emotion-guided multi-head attention that combines $\boldsymbol{f}_{t}^{'}$ and the output $\boldsymbol{E}_{e} \left( \boldsymbol{A}_{ci,ej} \right)$ of the emotion latent space is proposed. This module enhances the emotional expressiveness of 3D animated faces, as demonstrated by experiments conducted in this study (Tab.~\ref{table5}). Finally, $\boldsymbol{f}_{t}^{''}$ is fed into a feed-forward layer that outputs $\boldsymbol{f}_{t}^{'''}$, which is then passed through an audio-blendshape decoder implemented as a fully connected layer that outputs 52 blendshape coefficients.\par

\subsection{Loss function}
To train our neural network, we employ a loss function that comprises four distinct components: cross-reconstruction loss, self-reconstruction loss, velocity loss, and classification loss. The overall function is given by:
\begin{equation}\label{4}
L=\lambda_{1}L_{cross}+\lambda_{2}L_{self}+\lambda_{3}L_{vel}+\lambda_{4}L_{cls},
\end{equation}
where $\lambda_{1}$ = 1.0, $\lambda_{2}$ = 1.0, $\lambda_{3}$ = 0.5 and $\lambda_{4}$ = 0.1 in all of our experiments. We provide a detailed explanation of each of these components below.\par
\noindent\textbf{Cross-reconstruction loss.} In order to disentangle emotional content from speech signals, as described in Sec. \ref{section:cross}, we train our network to reconstruct various cross combinations and generate new blendshape coefficients. Given input audio $\boldsymbol{A}_{c1,e2}$ and $\boldsymbol{A}_{c2,e1}$, the encoder decomposes them and then reconstructs new combinations, which are compared with the ground truth blendshape coefficients $\boldsymbol{B}_{c1,e1}$ and $\boldsymbol{B}_{c2,e2}$. The formula is as follows:
\begin{equation}\label{5}
\begin{aligned}
L_{cross } & =\left\|\boldsymbol{D}\left(\boldsymbol{E}_{c}\left(\boldsymbol{A}_{c1, e2}\right), \boldsymbol{E}_{e}\left(\boldsymbol{A}_{c2, e1}\right)\right)-\boldsymbol{B}_{c1, e1}\right\|^{2} \\
& +\left\|\boldsymbol{D}\left(\boldsymbol{E}_{c}\left(\boldsymbol{A}_{c2, e1}\right), \boldsymbol{E}_{e}\left(\boldsymbol{A}_{c1, e2}\right)\right)-\boldsymbol{B}_{c2, e2}\right\|^{2} ,
\end{aligned}
\end{equation}
where D is the emotion-guided feature fusion decoder for reconstructing the cross combinations.

\noindent\textbf{Self-reconstruction loss.} While constraining the quality of the reconstructed output using cross-reconstruction, we also require the network to reconstruct its ground truth blendshape coefficients. The self-reconstruction loss can be expressed as:
\begin{equation}\label{6}
\begin{aligned}
L_{self} & =\left\|\boldsymbol{D}\left(\boldsymbol{E}_{c}\left(\boldsymbol{A}_{c1, e2}\right), \boldsymbol{E}_{e}\left(\boldsymbol{A}_{c1, e2}\right)\right)-\boldsymbol{B}_{c1, e2}\right\|^{2}.
\end{aligned}
\end{equation}\par

\noindent\textbf{Velocity loss.} To address the issue of jittery output frames when using only reconstruction loss, we introduce a velocity loss to induce temporal stability, which considers the smoothness of prediction and ground truth in the sequence context. By incorporating this loss, our model is encouraged to produce smoother and more realistic facial expressions. The velocity loss can be expressed as:
\begin{equation}\label{8}
\begin{aligned}
L_{vel}=\left\| \left(\boldsymbol{\hat{b}}_t - \boldsymbol{\hat{b}}_{t-1}\right)-\left(\boldsymbol{b}_t - \boldsymbol{b}_{t-1}\right)\right\|^{2},
\end{aligned}
\end{equation}\par

\noindent\textbf{Classification loss.} Due to the inherent difficulty of explicitly discerning the separability of emotional latent space during the disentangling process, we introduce a classification loss to supervise the output of the emotion extractor $\boldsymbol{E}_e$ and enhance its ability to discriminate between different emotions. The classification loss is defined as:
\begin{equation}\label{9}
\begin{aligned}
L_{cls}=-\sum_{i}\sum_{c=1}^{\boldsymbol{M}}\left(\boldsymbol{y}_{ic} * \log \boldsymbol{p}_{ic}\right),
\end{aligned}
\end{equation}
where $\boldsymbol{M}$ represents the number of distinct emotion categories, $\boldsymbol{y}_{ic}$ is the observation function that determines whether sample $i$ carries the emotion label $c$, and $\boldsymbol{p}_{ic}$ denotes the predicted probability that sample $i$ belongs to class $c$.
\subsection{Datasets construction}
\begin{figure}
    \centering
    \includegraphics[width=3.2in, keepaspectratio]{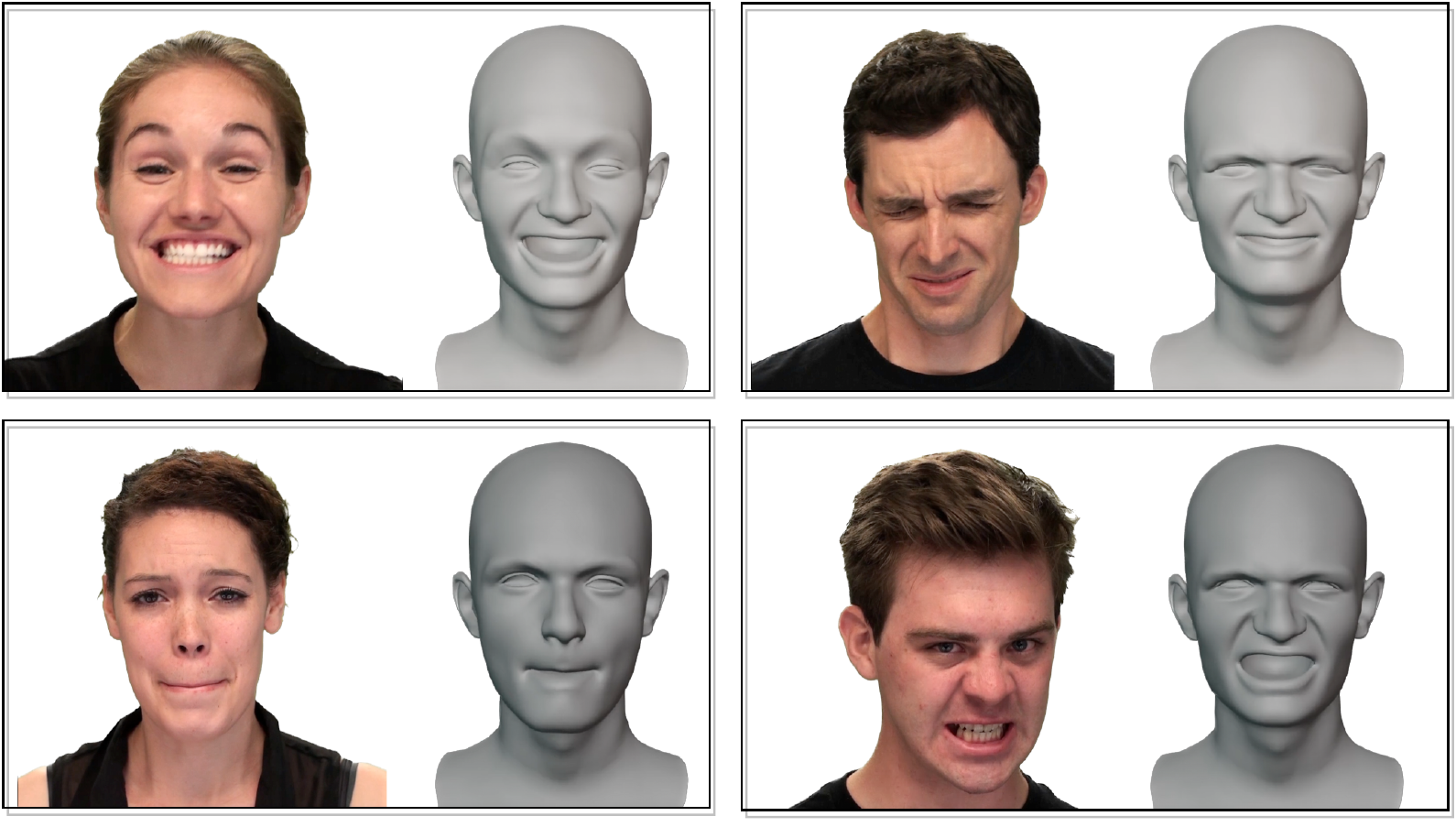}
    \caption{\textbf{Facial Blendshape Capture.} Input video streams of different expressions, and outputs the blendshape coefficients of the corresponding expressions.}
    \label{fig:4}
\end{figure}

Due to the scarcity of 3D talking face data with emotions, no such data is publicly available. To acquire such data, professional equipment and actors who can utter the same sentence with varied emotions are required, which entails high expenses. However, numerous 2D emotional audio-visual datasets exist. We employ facial blendshapes as a supervisory signal, which facilitates the reconstruction of plausible 3D faces from 2D images. Then, We extract blendshape coefficients from two datasets using a sophisticated blendshape capture method\footnote{Details can be found in the Supplementary Material.} which result accurately capture human emotional expressions (Fig.~\ref{fig:4}). A large 3D emotional talking face (3D-ETF) dataset consisting of approximately 700,000 frames of blendshape coefficients, spanning over 6.5 hours, is constructed using this method. Through blend linear skinning, both blendshape coefficients~\cite{lewis2014practice} and mesh vertices are built for the 3D-ETF dataset, filling a gap in 3D facial animation datasets, especially regarding emotional expression data and providing vivid and lifelike human facial expressions.

\section{Experiments}

\subsection{Datasets}
Two widely used 2D audio-visual datasets were utilized to construct the 3D-ETF dataset: RAVDESS~\cite{livingstone2018ryerson} and HDTF~\cite{zhang2021flow}.\par
The RAVDESS dataset~\cite{livingstone2018ryerson}, also known as the Ryerson Audio-Visual Database of Emotional Speech and Song, is a multi-modal emotion recognition dataset comprising 24 actors (12 male, 12 female) and 1440 video clips of short speeches. The dataset was captured with high-quality audio and video recordings, and the actors were instructed to express specific emotions, including neutral, calm, happy, sad, angry, fearful, disgusted, and surprised. A random selection of 80$\%$ of the dataset was used for training, 10$\%$ for validation, and 10$\%$ for testing.\par
The High-Definition Talking Face (HDTF) dataset~\cite{zhang2021flow} is a collection of approximately 16 hours of 720P-1080P videos sourced from YouTube over the past few years. The dataset includes over 300 subjects and 10k different sentences. Five hours of videos from the HDTF dataset were selected for mouth shape generalization and then partitioned into training, validation, and testing sets in the same proportion as the RAVDESS dataset.

\subsection{Quantitative evaluation}
To measure lip synchronization, we calculated the lip vertex error (LVE) as used in MeshTalk~\cite{richard2021MeshTalk} and FaceFormer~\cite{fan2022faceformer}. This evaluation metric computes the average $\ell_{2}$ error of the lips in the test set. For a single frame, LVE is defined as the maximum $\ell_{2}$ error among all lip vertices. Since LVE alone cannot reflect the full emotional expression, we proposed an emotional vertex error (EVE). To compute EVE, vertex indexes in the eye and forehead regions on the FLAME template are first selected. Similar to LVE, the EVE measures the maximum $\ell_{2}$ error of the vertex coordinate displacement in the interested region, and the average LVE over the test set is reported as the evaluation metric.\par 
\begin{table}[]
\begin{center}
    \tabcolsep=0.001cm
    \renewcommand\arraystretch{1.2}

\begin{tabular}{@{}lcccc@{}}
\toprule
              & \multicolumn{2}{c}{RAVDESS (emotion)}                                                                                                                   & \multicolumn{2}{c}{HDTF (no emotion)}                                                                                                                      \\ \midrule
Method        & \begin{tabular}[c]{@{}c@{}}LVE(mm)$\downarrow$ \end{tabular} & \begin{tabular}[c]{@{}c@{}}EVE(mm)$\downarrow$\\ \end{tabular} & \begin{tabular}[c]{@{}c@{}}LVE(mm)$\downarrow$\\ \end{tabular} & \begin{tabular}[c]{@{}c@{}}EVE(mm)$\downarrow$ \end{tabular} \\ \hline
VOCA~\cite{cudeiro2019capture}          & 5.091                                                                 & 4.188                                                                 & 4.447                                                                 & 3.286                                                                 \\
MeshTalk~\cite{richard2021MeshTalk}      & 3.459                                                                 & 3.386                                                                 & 3.886                                                                 & 3.124                                                                 \\
FaceFormer~\cite{fan2022faceformer}    & 3.247                                                                 & 3.757                                                                 & 3.374                                                                 & 3.142                                                                 \\
\textbf{Ours} & \textbf{2.762}                                                        & \textbf{2.493}                                                        & \textbf{2.892}                                                        & \textbf{2.364}                                                        \\ \bottomrule
\end{tabular}
\end{center}
\caption{\textbf{Quantitative evaluation results on RAVDESS and HDTF datasets.} The lip vertex error (LVE) and emotional vertex error (EVE) of our method are lower than those of the current state-of-the-art methods. }
\label{table1}
\end{table}
\begin{table}[]
\setlength\tabcolsep{8pt}
\begin{center}
    \renewcommand\arraystretch{1.2}
\begin{tabular}{lccc}
\toprule
Method                 & \begin{tabular}[c]{@{}c@{}}LVE(mm)$\downarrow$ \end{tabular} &  \begin{tabular}[c]{@{}c@{}}Train on VOCASET\end{tabular} \\ \midrule
VOCA~\cite{cudeiro2019capture}                      &   4.704                                                    & $\checkmark$                                        \\
MeshTalk~\cite{richard2021MeshTalk}                  &   4.513                                             & $\checkmark$                                        \\
FaceFormer~\cite{fan2022faceformer}               & 4.418                                                & $\checkmark$                                        \\
\textbf{Ours} & \textbf{4.134}                                 & \ding{55}                                      \\ \bottomrule
\end{tabular}
\end{center}
\caption{\textbf{Quantitative evaluation results on VOCA-Test.} Our method exhibits strong generalization capability in zero-shot cases and outperforms the current state-of-the-art methods.}
\label{table2}
\end{table}

\begin{table}[]
\begin{center}
\begin{tabular}{@{}lcccc@{}}
\toprule
Method     & \begin{tabular}[c]{@{}c@{}}RAVDESS\end{tabular} & \begin{tabular}[c]{@{}c@{}}HDTF\end{tabular} & \begin{tabular}[c]{@{}c@{}}MEAD\end{tabular} &\begin{tabular}[c]{@{}c@{}}VOCASET\end{tabular} \\ \midrule
VOCA       & 2.700                                                       & 2.427                &2.236                                       & 2.292                                                      \\
MeshTalk   & 2.139                                                       & 1.868                &2.058                                       & 2.070                                                      \\
FaceFormer & 1.958                                                       & 1.391                &1.852                                       & 1.944                                                      \\
\textbf{Ours}       & \textbf{1.648}                                              & \textbf{0.626}                & \textbf{1.498}                             & \begin{tabular}[c]{@{}c@{}}\textbf{1.914}\vspace{-0.5em}\\ {\footnotesize (zero-shot)}\vspace{-0.4em}\end{tabular}                                             \\ \bottomrule
\end{tabular}
\end{center}
\caption{\textbf{Quantitative evaluation results of lip average $\ell_{2}$ error.}}
\label{table3}
\vspace{-1em}
\end{table}
\begin{figure*}[t]
\begin{center}
   \includegraphics[width=0.93\linewidth]{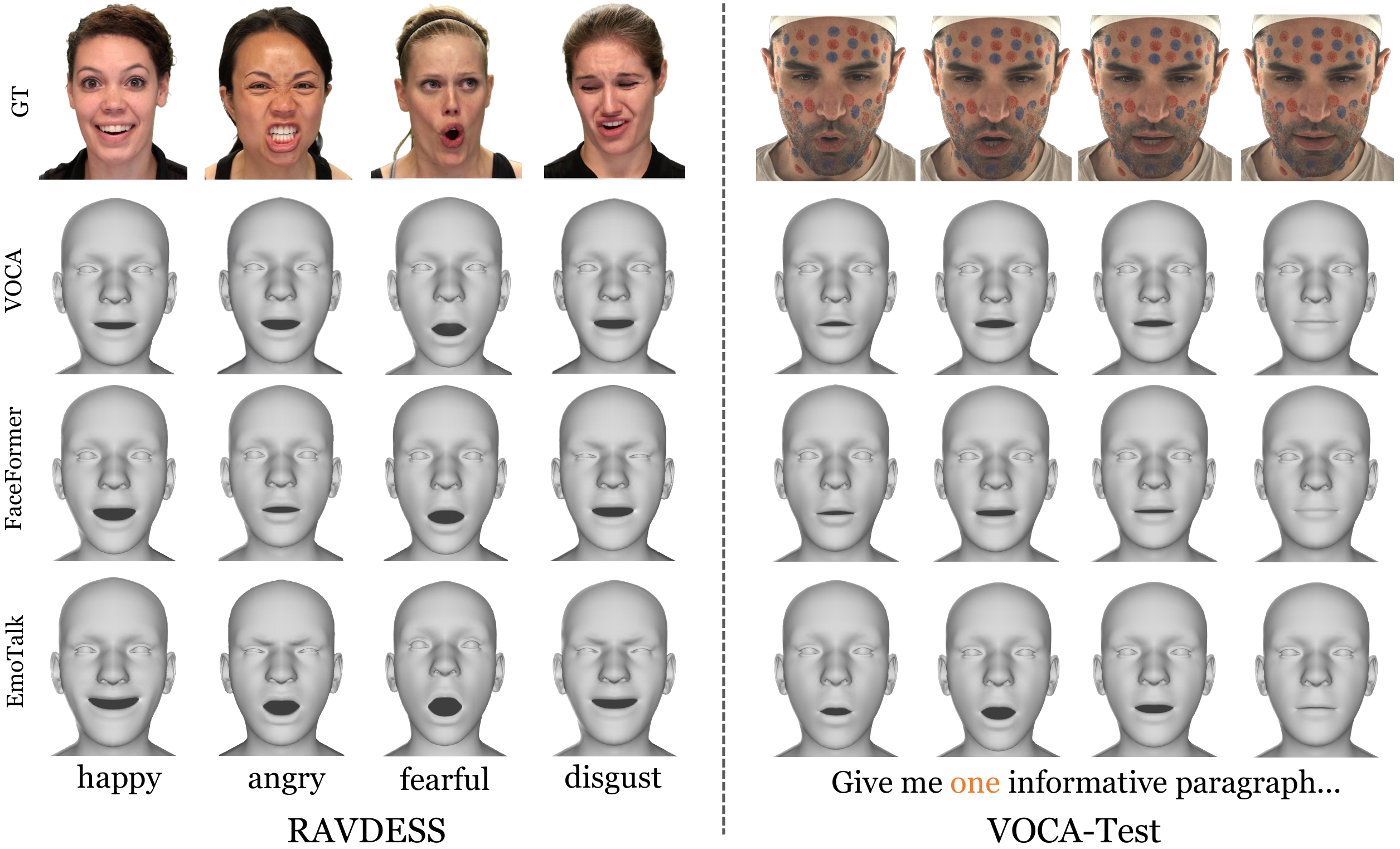}
\end{center}
   \caption{\textbf{Qualitative comparison of facial movement by different methods on RAVDESS (left) and VOCA-Test (right).} On RAVDESS, we generate facial animations of saying the word ``dog'' with different emotions. Our method can produce expressive facial movements that match the emotions. On VOCA-Test, we generate facial animations of saying the word ``one'' without emotion. Our method can achieve similar performance to the ground truth, and the range of motion is noticeable. }
\label{fig:5}
\vspace{-0.2in}
\end{figure*}
We retrained VOCA, MeshTalk, FaceFormer, and our method (EmoTalk) on the RAVDESS and HDTF datasets. The blendshape coefficients were converted into mesh vertices (5023*3) corresponding to the FLAME model, which was used as ground truth. Tab.~\ref{table1} shows LVE and EVE evaluation results. EmoTalk achieved lower lip error and emotion expression error than the three previous methods.The proposed model has more accurate lip movements and better emotional expression.\par
\noindent\textbf{Generalization analysis.} 
The model could not be trained on VOCASET because access to the corresponding blendshape coefficients was unavailable, and the blendshape capture method was incompatible with the marked facial images provided by the official dataset. Nevertheless, we evaluated our model on this dataset by converting the output blendshape coefficients to mesh vertices and comparing them with ground truth.\par
As reported in Tab.~\ref{table2}, EmoTalk outperformed the other methods on VOCA-Test, even in zero-shot settings. This could be attributed to two reasons: (i) controlling facial animation through blendshape coefficients has a higher generalization ability than predicting vertex offsets based on mesh, and (ii) sufficient 2D datasets can also enable the model to learn complex relations between speech and facial expressions, thus achieving better results. 

\noindent\textbf{Robustness analysis.} When using the lip maximum $\ell_{2}$ error metric, there may be a potential impact of outliers present in the dataset. To mitigate the impact of outliers and present a more comprehensive evaluation, we additionally computed the lip average $\ell_{2}$ error for proposed method and introduced the MEAD dataset~\cite{wang2020mead}. In Tab.~\ref{table3}, we present the results of the lip average $\ell_{2}$ errors obtained from our method and previous methods. The analysis of these errors demonstrates the superior performance of our proposed method over multiple datasets, substantiating its effectiveness and robustness in comparison to existing methods.

\subsection{Qualitative evaluation}

As audio and facial movements cannot be evaluated solely based on indicators and require human perceptual evaluation, we conducted a qualitative assessment of our model from two perspectives.\par
\noindent\textbf{Lip synchronization.} We compared our model with VOCA and FaceFormer by feeding them identical audio inputs and generating corresponding facial animations. The results showed that the proposed model exhibited more pronounced lip movements and better alignment with human speech patterns. Even in cases of rapid mouth movements, such as pronouncing the word ``shy", where the lips should gradually open and then close, the proposed model captured this lip synchronization more effectively (Fig. \ref{fig:5}). Moreover, the proposed model’s ability to close the mouth was comparable to the results trained on high-precision scanned facial datasets.\par
\noindent\textbf{Emotional expression.} Previous methods were not optimized for emotional expression, resulting in limited facial expressions for different speech patterns. However, our method could clearly demonstrate variations among emotions. For example, in anger emotion, the movements of raising and lowering eyebrows could distinctly reflect the emotional information conveyed by speech (Fig. \ref{fig:5}). A supplementary video provides more detailed comparisons.

\subsection{User study}
To evaluate the proposed model more thoroughly, We designed a comprehensive user questionnaire like~\cite{richard2021MeshTalk} and performed a comparative analysis with MeshTalk~\cite{richard2021MeshTalk} and FaceFormer~\cite{fan2022faceformer} using the FLAME template. Since our model incorporates emotional components, we devised three sub-tasks: full-face comparison, lip synchronization comparison (by covering the area above the nose), and emotion expression comparison (by covering the area below the nose). We selected twenty sentences from the RAVDESS and VOCASET test datasets as our test cases, ultimately formulating 120 multiple-choice questions. The questionnaire system randomly presented a pair of comparison videos to users, allowing them to choose which video is more realistic. We counted how many users chose our result versus the competitor's result, and the ratio of user choice was calculated for satisfaction evaluation. \par 
Specifically, compared with MeshTalk and FaceFormer, our model received the most positive feedback from participants, surpassing MeshTalk and FaceFormer in full-face voting by 65.9$\%$ and 64.6$\%$, respectively. Notably, in terms of emotional expression, our model displays a huge advantage over alternative approaches. Overall, most participants considered our method superior to MeshTalk and FaceFormer. Detailed user choice results are shown in Tab.~\ref{table4}.

\subsection{Ablation experiment}



We conducted an ablation study to examine the contributions of different components of our model. The essential modules, loss functions, and datasets were individually studied to examine their effects on the evaluation metrics. In Tab.~\ref{table5}, it is shown that a significant increase in emotional expression error ensued when removing the emotion disentangling encoder, which demonstrated the critical role of EDE in emotional learning and expression. Similarly, removing the emotion-guided multi-head attention also witnessed a certain increase in EVE, which indicated the effectiveness of the emotion guidance module in enhancing emotional expression.\par
From the loss function aspect, removing velocity loss leads to a slight drop in performance, but it caused noticeable jitter in the output of facial animation. Removing classification loss clearly increased the EVE, suggesting that the feature extractor could distinguish emotions less effectively. Then we train our model without using the HDTF dataset to investigate the LVE changes. It is observed that the LVE increases by about $0.5mm$, and the performance of lip vertex prediction decayed dramatically. This indicates that training with the HDTF dataset is able to learn more about mapping relationships between lip movements and speech. Finally, we replaced our emotion disentangling encoder with the method proposed by Ji et al~\cite{ji2021audio}. because they used MFCC as the audio feature extractor, and their approach required staged training during the processing. As a result, we observed an increase in errors during the evaluations of LVE and EVE. This indicates the effectiveness of the improvements we made on the Ji et al.'s method~\cite{ji2021audio}.\par

\begin{table}[]
\setlength\tabcolsep{11pt}
\begin{center}
\begin{tabular}{lcc}
\toprule
Method                       & Competitors      & Ours            \\ \hline
\textbf{Ours vs. MeshTalk}   &                 &                 \\
full-face                    & 34.1\%          & \textbf{65.9\%} \\
lip sync                     & 38.7\%          & \textbf{61.3\%} \\
emotion expression           & 31.5\%          & \textbf{68.5\%} \\ \midrule
\textbf{Ours vs. FaceFormer} &                 &                 \\
full-face                    & 35.4\%          & \textbf{64.6\%} \\
lip sync                     & 40.9\%          & \textbf{59.1\%} \\
emotion expression           & 30.8\%          & \textbf{69.2\%} \\ \bottomrule
\end{tabular}
\end{center}
\caption{\textbf{User study results.} We devised three sub-tasks, namely full-face comparison, lip synchronization comparison, and emotion expression comparison.}
\label{table4}
\end{table}
\begin{table}[]
\setlength\tabcolsep{9pt}
\begin{center}
\begin{tabular}{@{}lcc@{}}
\toprule
& \begin{tabular}[c]{@{}c@{}}LVE (mm)\end{tabular} & \begin{tabular}[c]{@{}c@{}}EVE (mm)\end{tabular} \\ \midrule
\textbf{Ours }                                                                                       &\textbf{2.762}     &\textbf{2.493}     \\ \midrule
\begin{tabular}[c]{@{}l@{}}w/o Emotion Disentangling \\ Encoder\end{tabular} & 3.126    & 3.076    \\
\begin{tabular}[c]{@{}l@{}}w/o Emotion-Guided \\Multi-Head Attention\end{tabular}        & 2.907    & 2.832    \\ \midrule
w/o $L_{vel}$ Loss                                                                                 & 2.813    & 2.775    \\
w/o $L_{cls}$ Loss                                                                              & 3.096    &  2.815   \\ \midrule
w/o HDTF Dataset                                                                            & 3.254    & 2.806    \\
\begin{tabular}[c]{@{}l@{}}replace our Encoder \\with Ji et al.'s~\cite{ji2021audio}\end{tabular}                 & 3.583    & 2.973    \\\bottomrule
\end{tabular}
\end{center}
\caption{\textbf{Ablation study for our components.} We show the LVE and EVE in different cases.}
\label{table5}
\end{table}
\section{Limitations}

 Our method still has some limitations we plan to address in future work. First, our method relies on a large-scale audio pre-training model, which increases the inference time and hinders real-time applications. Second, our network outputs 52 blendshape coefficients, which do not include head movements, \eg head shakes and rotations. A possible solution is to combine blendshape coefficients with the FLAME model~\cite{li2017learning} to control both facial expressions and head movements. Third, our training data is derived from 2D images. The pseudo-3D data is not as precise as 3D-scanned data and thus cannot represent the skin's micro facial expressions. As a result, our method can only reflect the overall emotional state of the animated face. We intend to collect more emotional data using professional instruments in the future and share it with the research community.\par
\section{Conclusion}
This paper proposes a novel method for generating speech-driven 3D face animation that effectively conveys emotions. Our method consists of two key components: an emotion disentangling encoder and an emotion-guided feature fusion decoder. The former segregates the speech into its emotional and content components, providing clear emotional information for facial animation. The latter enhances the expressiveness of facial animation by emphasizing emotion-related features. To address the problem of missing 3D emotional talking face data, we construct a large-scale 3D emotional talking face (3D-ETF) dataset that contains blendshape coefficients and mesh vertices.
Additionally, we have implemented parameterized transformations for blendshape coefficients and the FLAME model, allowing for efficient conversion between various facial animations. Experimental results demonstrate that our method outperforms existing state-of-the-art methods and receives better user experience feedback. Our work contributes to virtual reality applications. It can enable more realistic and immersive virtual experiences with emotional talking faces.

\section*{Acknowledgments}
This work was supported in part by National Key Research and Development Program of China under Grant No. 2020YFB2104101 and National Natural Science Foundation of China (NSFC) under Grant Nos. 62172421, 62072459 and 71771131.

{\small
\bibliographystyle{ieee_fullname}
\bibliography{egbib}
}

\clearpage
\section*{Appendix}
\begin{appendices}
In this supplementary material, we provide more details about EmoTalk, which consists of five parts: 1) The implementation details of EmoTalk, including the model architecture and parameter details; 2) The transform module from blendshape to FLAME head, including the transform method and calculation formula; 3) The comparison method with baselines, including the comparison objects and evaluation details; 4) The construction details of the 3D-ETF dataset, including data collection, preprocessing, and post-processing; 5) The implementation details of blendshape capture method.
\section{Implementation details}
EmoTalk’s overall architecture is illustrated in Fig. 2 of the main paper. In order to improve the reproducibility and credibility of EmoTalk on the 3D emotional face animation generation task, we will further explain how we design and implement two key components: emotion disentangling encoder and emotion-guided feature fusion decoder.
\subsection{Training details}
The network receives preprocessed video and audio data as input. The video stream is converted to 30 frames per second, while the audio sampling rate is 16 kHz. A facial blendshape capture method generates facial parameters consisting of 52 blendshape coefficients per frame for the video data.\par
During the training process, the model is optimized end-to-end using the Adam optimizer~\cite{Kingma2014AdamAM}. The learning rate and batch size are set to 1$e$ - 4 and 8, respectively. The model is trained on a single NVIDIA V100, and the entire network takes approximately 8 hours (80 epochs) to train.
\subsection{Emotion disentangling encoder}
To perform emotion disentanglement, we first convert the input audio signal to a sampling rate of 16 KHz. Then we encode it using temporal convolutional network (TCN) to process sequential data with convolutional architecture. Next, we use a linear interpolation layer to adjust the length of the encoded representation according to the target audio signal. For instance, if we want to reconstruct $\boldsymbol{A}_{c1,e1}$ using $\boldsymbol{A}_{c1,e2}$ and $\boldsymbol{A}_{c2,e1}$ as inputs, then we need to interpolate them to have the same length as $\boldsymbol{A}_{c1,e1}$. After that, we decode the interpolated representation using 24 transformer\cite{vaswani2017attention} blocks. Each transformer block has a model dimension of 1024, an inner dimension of 4096, and 16 attention heads. Finally, we obtain two feature vectors of dimension 1024 each, representing content and emotional information in the output audio signal from pre-trained models. 
We use a cross-reconstruction constraint method to optimize model parameters during the training process, which we detail in Sec 3.1 of the main paper.
\subsection{Emotion-guided feature fusion decoder}
We first map the output of the features by the emotion feature extractor and the content feature extractor to 256-dimensional and 512-dimensional vectors, respectively. Then we add two one-hot embeddings for emotion level and personal style, each mapped to a 32-dimensional vector. The emotion level is a binary variable indicating high or low intensity, while the personal style is a multivariate variable representing 24 different speakers. We concatenate these four features to form an 832-dimensional feature vector. We also add a periodic position encoding\cite{fan2022faceformer} of the same dimension to this vector. 
Moreover, we use a fully connected layer to reduce the dimension of the output of the features by the emotion encoder from 1024 to 832 for subsequent emotion guidance. For biased multi-head self-attention and emotion-guided multi-head attention, we use four heads and set the dimension to 832 for each transformer decoder block. The concatenated features serve as the input sequence for the decoder, while emotional features serve as the output sequence from the last encoder layer, thus achieving emotion guidance. Finally, we feed the forward layer’s output into the audio-blendshape decoder, which is a fully connected layer that maps between 832 dimensions and 52 dimensions blendshape coefficients. Thus we obtain emotion-enhanced blendshape coefficients.
\section{Blendshape to FLAME transform module}
\begin{figure*}
\begin{center}
    \includegraphics[width=0.9\textwidth, keepaspectratio]{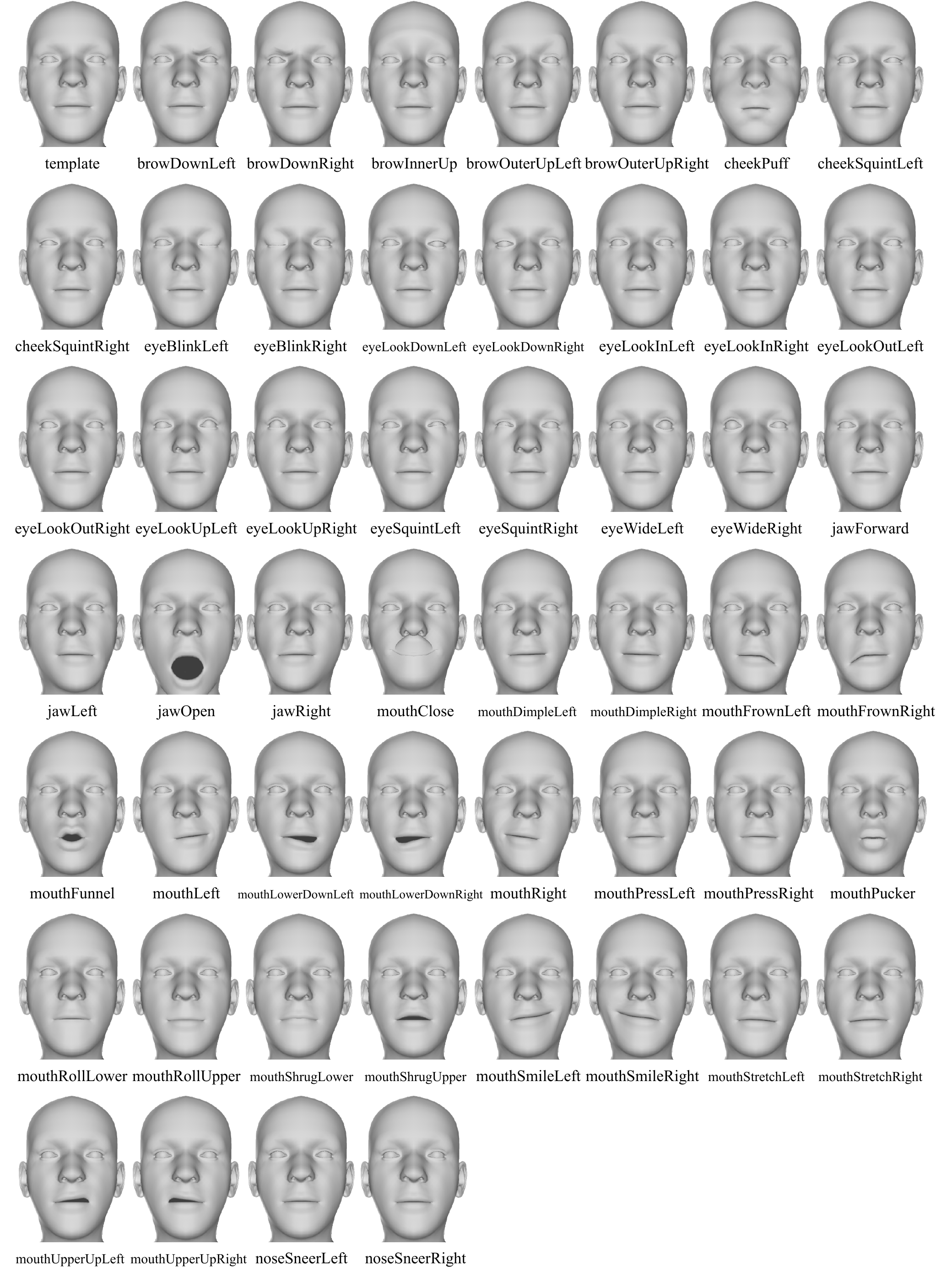}
\end{center}
   \caption{\textbf{Semantically Meaningful FLAME Head Templates.} We create 52 FLAME head templates that correspond to the blendshape coefficients, to achieve the transformation from the blendshape coefficients to the FLAME head model.}
\label{fig:7}
\end{figure*}

The Blendshape\cite{lewis2014practice} to FLAME\cite{li2017learning} transform module converts blendshapes, which is a way of deforming a mesh by interpolating between different shapes, to a FLAME head, which is a 3D head model that captures variations in identity, expression, head pose and gaze. This transform module enables our model to transfer facial expressions across different virtual characters quickly. To achieve this conversion, we collaborated with professional animators to create 52 semantically meaningful FLAME head templates (see Fig.~\ref{fig:7}). These templates allow us to obtain the facial deformation parameters corresponding to blendshape and mesh head. We use blend linear skinning to interpolate between these parameters. Because blendshape labels have semantic meanings, they can quickly transfer facial motions across different virtual characters.

Specifically, after obtaining the blendshape coefficients output by EmoTalk, we perform linear weighting on the corresponding parameters of 52 FLAME head templates to obtain the vertex parameters of 5023*3 dimensions. The formula is as follows:
\begin{equation}\label{1}
V_{flame} = \sum_{i=1}^{52} \beta V_i
\end{equation}\par

where $\boldsymbol{V}_{flame}$ is the final output of FLAME head vertex coordinates, $\boldsymbol{V}_i$ is the vertex coordinate of the $\boldsymbol{i}^{th}$ FLAME head template, and $\boldsymbol{\beta}$ is the blendshape coefficient vector output by EmoTalk.
\section{Baseline methods}
We conducted a comparative analysis of EmoTalk with three state-of-the-art approaches, namely VOCA\cite{cudeiro2019capture}, MeshTalk\cite{richard2021MeshTalk}, and FaceFormer\cite{fan2022faceformer}. To facilitate a comprehensive evaluation, we employed two distinct datasets, namely the RAVDESS and HDTF, both of which are processed through our facial blendshape capturing technique to obtain the ground truth. For each frame in the datasets, we calculated the blendshape coefficients and mapped them to the corresponding vertex parameters of the FLAME model using the transform module. Furthermore, we retrained the models of the three existing approaches using RAVDESS , HDTF   and 3D-ETF datasets to improve their performance. \par 
For VOCASET, we used the pre-trained models provided by VOCA and FaceFormer and retrained the MeshTalk model to evaluate the vertex error of these three methods on the VOCA-Test. It is worth noting that due to the absence of blendshape coefficients in the official VOCASET dataset and the images containing marked faces incompatible with our blendshape capturing approach, we are unable to train our model on this dataset. Instead, we directly evaluated the EmoTalk model, trained on the HDTF dataset, on VOCA-Test. \par
During the evaluation, while the other three methods computed the error directly between the output vertices and the ground truth, we needed to use a transfer module to convert the EmoTalk output from blendshape coefficients to mesh vertices to ensure comparability with other methods in the same dimension and eliminate any differences between output formats.
\section{Dataset construction details}
In this study, we constructed a large 3D emotional talking face (3D-ETF) dataset, where facial blendshape is used as the supervisory signal to reconstruct reliable 3D faces from 2D images. The facial blendshape capturing method is fine-tuned by animators to create numerous 3D facial animations from the RAVDESS\cite{livingstone2018ryerson} and HDTF\cite{zhang2021flow} datasets.\par
Specifically, 1440 videos from the RAVDESS dataset and 385 videos from the HDTF dataset are processed by converting them into 30 frames per second and capturing the facial blendshape for each frame. To enhance the quality of the dataset and reduce frame-to-frame jitter, a Savitzky-Golay filter with a window length of 5 and a polynomial order of 2 is applied to the output blendshape coefficients, which significantly improved the smoothness of facial animation. The RAVDESS dataset generated 159,702 frames of blendshape coefficients, which amounts to approximately 1.5 hours of video content. Meanwhile, the HDTF dataset generated 543,240 frames of blendshape coefficients, which equates to approximately 5 hours of video content. All the generated blendshape coefficients are converted into mesh vertices using the transform module and included in the dataset. A supplementary video will demonstrate the effectiveness of our dataset.\par
\section{Blendshape capture method}
Our sophisticated blendshape capture method predicts corresponding blendshape coefficients from input video streams using a neural network model, which is then manually fine-tuned by professional animators to achieve realistic facial reconstruction results that accurately capture human emotional expressions. \par
In this method, we use the ``Live Link Face" application to collect  a dataset consisting of images paired with corresponding blendshape data. The image preprocessing involved facial cropping and other necessary transformations before feeding them into a ResNet~\cite{he2016deep} architecture. The ResNet model was employed to produce 52 specific blendshape values as the output, and these values were constrained using the L2 loss function, ensuring precise regression of facial blendshapes.
\end{appendices}
\end{document}